# Corpus-Driven Knowledge Acquisition for Discourse Analysis

Stephen Soderland and Wendy Lehnert [*]
Department of Computer Science
University of Massachusetts
Amherst, MA 01003-4610
soderlan@cs.umass.edu  lehnert@cs.umass.edu

## Abstract

The availability of large on-line text corpora provides a natural and promising bridge between the worlds of natural language processing (NLP) and machine learning (ML). In recent years, the NLP community has been aggressively investigating statistical techniques to drive part-of-speech taggers, but application-specific text corpora can be used to drive knowledge acquisition at much higher levels as well. In this paper we will show how ML techniques can be used to support knowledge acquisition for information extraction systems. It is often very difficult to specify an explicit domain model for many information extraction applications, and it is always labor intensive to implement hand-coded heuristics for each new domain. We have discovered that it is nevertheless possible to use ML algorithms in order to capture knowledge that is only implicitly present in a representative text corpus. Our work addresses issues traditionally associated with discourse analysis and intersentential inference generation, and demonstrates the utility of ML algorithms at this higher level of language analysis.

The benefits of our work address the portability and scalability of information extraction (IE) technologies. When hand-coded heuristics are used to manage discourse analysis in an information extraction system, months of programming effort are easily needed to port a successful IE system to a new domain. We will show how ML algorithms can reduce this development time to a few days of automated corpus analysis without any resulting degradation of overall system performance.

## 1. Information Extraction at the Discourse Level

All IE systems must operate at both the sentence level and the discourse level. At the sentence level, relevant information is extracted by a sentence analyzer according to pre-defined domain guidelines. Recent performance evaluations sponsored by ARPA have shown that a number of different parsing strategies can handle sentence-level information extraction with varying degrees of success (Lehnert and Sundheim 1991, Sundheim 1991).

This paper will concentrate on the discourse level, by which we mean all processing that takes place after sentence analysis. Once information has been extracted locally from various text segments, the IE system must make a series of higher-level decisions before producing its final output. Multiple referents must be merged when they are coreferent, important relationships between distinct referents must be recognized, and referents that are spurious with respect to the IE application must be discarded.

To get a sense of the decisions involved in discourse, consider the following fragment of a text from the MUC-5 micro-electronics domain.

```
GCA unveiled its new XLS stepper, which was
developed with assistance from Sematech.  The
system will be available in deep-ultraviolet
and I-line configurations.
```

Sentence analysis should extract two company names and a piece of stepper equipment from the first sentence and two processes, UV lithography and I-line lithography, from the second sentence. It is up to discourse analysis to determine the relationships between these objects.

Domain guidelines require pointers in the output from a micro-chip fabrication processes to related equipment, from equipment to its manufacturer, and from a process to devices produced. There are four possible links between company and process: developer, manufacturer, distributor, and purchaser/user. Considerable domain knowledge is needed at the discourse-level to recognize these various relationships.

Wrap-Up is an ML-based discourse component for IE applications that automatically derives this domain knowledge from a training corpus and requires no hand-coded domain knowledge. It uses a series of interacting decision trees to make decisions about merging, linking, splitting, and discarding information produced by a sentence analyzer. During its training phase, Wrap-Up repeatedly consults an output key associated with each training text in order to construct decision trees that are later used to guide decisions when Wrap-Up operates as a stand-alone discourse analyzer.

---

[*]This research was supported by NSF Grant no. EEC-9209623, State/Industry/University Cooperative Research on Intelligent Information Retrieval.

## 2. Applying Decision Tree Algorithms to the Problem

Wrap-Up breaks discourse processing into a number of small decisions and builds a separate ID3 decision tree for each (Quinlan 1986). The Lithography-Equipment-Links tree is typical of the 91 decision trees used for the micro-electronics domain. During discourse processing, Wrap-Up encodes an instance for each pair of extracted lithography and equipment objects, such as UV lithography and XLS stepper in the previous example. If the Lithography-Equipment-Links tree returns a classification of "positive", a pointer is added from the lithography process to the equipment.

Much of the art of machine learning is in choosing suitable features for the instances. Wrap-Up's goal is to supply ID3 with all the information available from the sentence analyzer, but to avoid any domain-specific feature generators. Extraction of UV lithography was triggered by the linguistic pattern "available in X" and by the keyword "deep-ultraviolet". Wrap-Up encodes these as the binary features pp-available, pp-in, and keyword-deep-ultraviolet. The feature trigger-count has a value of 3.

The same is done for the linguistic context of Stepper, which was found in "unveils X", "X was developed", and by the keyword "stepper". The relative position of the two objects is captured by features for the number of common-phrases, the common-triggers, and the relative distance, which is -1 sentences apart.

```
(lithography-type . uv) (trigger-count-1 . 3)
(pp-1-available . t) (pp-1-in . t)
(keyword-1-deep-ultraviolet . t)
(equipment-type . stepper) (equipment-name . t)
(trigger-count-2 . 3) (dir-obj-2-unveiled . t)
(subj-passive-developed . t) (keyword-2-stepper . t)
(common-triggers . 0) (common-phrases . 0)
(distance . -1)
```

During the training phase, ID3 is given such an instance for every pair of lithography and equipment objects in the 800 training texts. If the hand-coded output key for the training text has a link between the lithography and equipment objects, the training instance is classified as positive. ID3 tabulates how often each possible feature value is associated with a positive or negative training instance and encapsulates these statistics at each node of the tree it builds.

As figure 1 shows, the Lithography-Equipment-Links tree started with 282 positive and 539 negative training instances, giving a 34% a priori probability of a link. ID3 recursively selects features to partition the training instances according to an information gain metric (p.89-90 Quinlan 1986). The feature chosen as root of this tree is equipment-type. This feature alone is sufficient to classify instances with equipment-type such as modular-equipment, radiation-source, or etching-system, which have only negative instances. Equipment-type stepper has 202 positive and 174 negative training instances, raising the probability of a link to 54%.

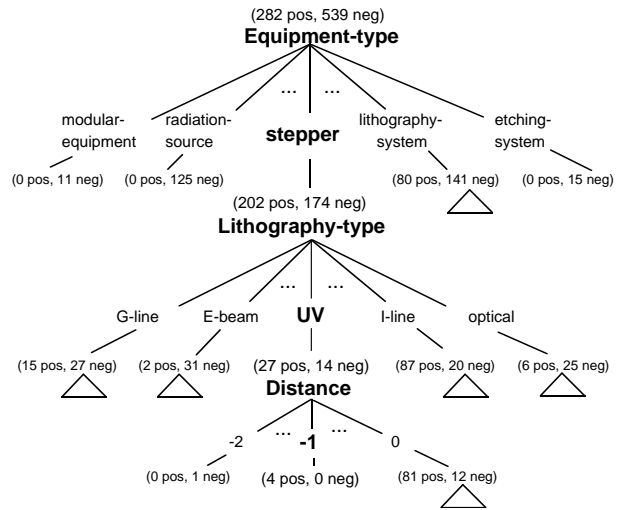

Figure 1: A Lithography-Equipment-Links decision tree. The highlighted path is for an instance with stepper equipment, UV lithography, and equipment mentioned one sentence earlier than lithography.

The next feature selected is lithography-type. The partition for UV lithography has 27 positive and 14 negative instances in contrast to e-beam, which has 94% negative instances. The next test is distance, with the branch for -1 leading to a leaf node with 4 positive and no negative instances. The tree returns a classification of positive and Wrap-Up adds a link from UV lithography to the Stepper.

This example shows how a decision tree can acquire useful domain knowledge: that lithography is never linked to equipment such as etching-system, and that steppers are often linked with UV lithography but hardly ever with e-beam lithography.

Other decision trees make greater use of the linguistic pattern features. Tests for linguistic patterns such as "X unveiled stepper" are used in the decision tree that filters out irrelevant company names and in the trees that decide whether a company is a developer of a micro-electronics process.

Using specific linguistic patterns resulted in extremely large, sparse feature sets for most trees. The Lithography-Equipment tree had 1045 features, all but 11 of them encoding linguistic patterns. Since each instance participates in at most a dozen linguistic patterns, a potential time and space bottleneck could be

> The Semiconductor Division of Mitsubishi Electronics America, Inc. now offers 1M CMOS DRAMs in Thin Small-Outline Packaging (TSOP*), providing the highest memory density available in the industry. Developed by Mitsubishi, the TSOP also lets designers increase system memory density with standard and reverse, or "mirror image," pin-outs. Mitsubishi's 1M DRAM TSOP provides the density to be 100% burned-in and fully tested. *Previously referred to as VSOP (very small-outline package) or USOP (ultra small-outline package). The 1M DRAM TSOP has a height of 1.2 mm, a plane measurement of 16.0 mm x 6.0 mm, and a lead pitch of 0.5 mm, making it nearly three times thinner and four times smaller in volume than the 1M DRAM SOJ package. The SOJ has a height of 3.45 mm, a plane dimension of 17.15 mm x 8.45 mm, and a lead pitch of 1.27 mm. Additionally, the TSOP weighs only 0.22 grams, in contrast with the 0.75 gram weight of the SOJ.
> Full text available on PTS New Product Announcements.

Figure 2: A sample text from the MUC-5 Microelectronics domain. Extracted information is underlined.

avoided by a sparse-vector implementation of ID3.

Tree pruning was also used when partitions near the leaf nodes become so small that the features selected have little predictive power. Wrap-Up empirically sets pruning level and threshold for each tree. The threshold determines the classification when a tree probe halts at a node with both positive and negative instances.

## 3. Wrap-Up: An Overview

Wrap-Up is a discourse component for information extraction that uses a series of interacting ID3 decision trees to make decisions about merging, linking, splitting, and discarding locally extracted information. The number of decision trees depends on the number of objects and links defined in the output structure. The feature set for each tree is automatically derived from linguistic patterns that occur in training instances. Wrap-Up provides a domain-independent framework which is instantiated for each domain with no additional heuristics or hand-coded knowledge needed.

Input to Wrap-Up is a set of tokens, each initially representing a single referent identified by the sentence analyzer. Tokens consist of a case frame containing the extracted information and a list of references to that information in the text with the location of each reference and the linguistic patterns used to extract it. Wrap-Up transforms this set of tokens, discarding information judged irrelevant to the domain, merging tokens with related information, adding pointers between tokens, and adding inferred tokens and default slot values.

Wrap-Up was tested using output extracted by the University of Massachusetts CIRCUS sentence analyzer (Lehnert 1990, Lehnert et al. 1992a, 1992b), although it could be adapted to any sentence analyzer which uses linguistic patterns for extraction.

Wrap-Up has six stages of processing, each with its own set of decision trees to guide the transformation of tokens as they are passed from one stage to the next.

Algorithm:

1. Slot Filter

Each token slot has its own decision tree that judges whether the slot contains reliable information. Discard the slot from a token if a tree returns "negative".

2. Slot Merge

Create an instance for each pair of tokens of the same type. Merge the two tokens if a decision tree for that token type returns "positive".

3. Links

Beginning at the lowest level of links in the output structure, consider pairs of tokens which might possibly be linked. Add a pointer between tokens if a decision tree returns "positive".

4. Links Merge

During the Links stage, token A may have a link to both token B and to token C. If a links-merge tree returns "positive", add pointers from token A to both B and C. If the tree returns "negative", split A into two copies with one pointing to B and the other to C.

5. Orphans

Orphans are tokens not pointed to by any other token. A decision tree returns the most likely parent token for each orphan. Create such a parent and link it to the orphan unless the tree returns "none". Then use decision trees from the Links and Links Merge stages to tie the new parent in with other tokens.

6. Slot Defaults

Create an instance for each empty token slot with a closed class of possible values. Add the slot value returned by a decision tree unless "none" is returned.

Perhaps the best way to understand the algorithm is to look at a concrete example. Figure 2 has a sample microelectronics text about packaging processes used to manufacture DRAM chips. The target output has SOJ packaging, TSOP packaging, one entity (Mitsubishi Electronics America, Inc), and a DRAM device. The size 1 MBit should be merged with DRAM and the material plastic merged with TSOP but not with SOJ packaging. Mitsubishi is linked to SOJ packaging as purchaser/user and to TSOP packaging as both developer and purchaser/user.

The first stage of Wrap-Up considers each slot of each extracted object to filter out irrelevant or spurious information. This step was included because the output of the sentence analyzer often includes spurious information. In this case the sentence analyzer correctly extracted "Mitsubishi Electronics America, Inc." but also reported a separate entity, "Semiconductor Division of Mitsubishi Electronics America, Inc.".

The Entity-Name-Filter tree was able to classify the former as a positive instance and the latter as negative. The first feature tested by this tree is trigger-count, since it turns out that entity names extracted by five linguistic triggers are more reliable than those extracted by only two triggers. This first test raises the confidence in "Mitsubishi Electronics America, Inc." with five linguistic triggers to 67% while lowering the confidence in "Semiconductor Division" to 36%. The tree then tested for various linguistic-triggers such as pp-of, subj-announced, pp-sold, and pp-subsidiary to arrive at a classification. A Packaging-Material-Filter tree also discarded epoxy, which was often spuriously extracted from training texts.

The next stage of Wrap-Up is slot-merge, where plastic is merged with TSOP packaging, but not with the SOJ packaging. A separate instances is created for each pair of packaging objects: TSOP-plastic, SOJ-plastic, and TSOP-SOJ. The Packaging-Slotmerge tree classifies TSOP-plastic as positive, since the distance is 0 and this combination occurs often in the training corpus. SOJ-plastic is classified negative, primarily because the distance is -2. TSOP-SOJ is easily classified as negative, since objects in the training instances never have multiple packaging types. The state of the output at this point in discourse processing is shown in figure 3.

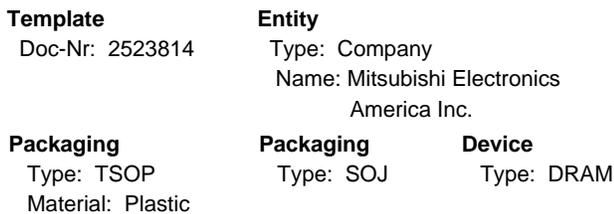

**Template**
Doc-Nr: 2523814

**Entity**
Type: Company
Name: Mitsubishi Electronics
America Inc.

**Packaging**
Type: TSOP
Material: Plastic

**Packaging**
Type: SOJ

**Device**
Type: DRAM

Figure 3: Output from the sample text before links have been added. One spurious company name and one packaging material have been filtered out and TSOP packaging has been merged with the packaging-material plastic.

Much of Wrap-Up's work occurs during the links stage, which first consults a Packaging-Device-Links tree to determine links between DRAM and each pack-

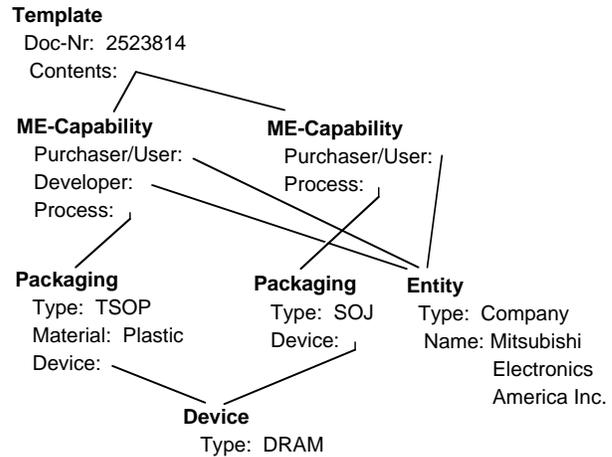

**Template**
Doc-Nr: 2523814
Contents:

**ME-Capability**
Purchaser/User:
Developer:
Process:

**ME-Capability**
Purchaser/User:
Process:

**Packaging**
Type: TSOP
Material: Plastic
Device:

**Packaging**
Type: SOJ
Device:

**Entity**
Type: Company
Name: Mitsubishi
Electronics
America Inc.

**Device**
Type: DRAM

Figure 4: Final output after links have been added

aging process. The root of this tree is the feature distance, followed by a test for packaging-type. Although only 29% of the training instances were positive, those with distance 0 and packaging-type TSOP were 80% positive. The SOJ-DRAM instance, with a distance of 0 was also classified positive.

The most difficult discourse decision is resolving the role of each company as developer, manufacturer, distributor, or purchaser/user of a process. Linguistic patterns such as "X is shipping", "X purchased", or "X developed" are important in distinguishing the company's role. But a company that "developed a new chip" is probably not developer of the process to fabricate that chip.

There were seldom explicit linguistic clues about the relation of a company to a packaging process, so trees such as Packaging-Purchaser-Links fell back on statistics based on relative distance in the text, packaging-type, and trigger-count. Although SOJ is first mentioned two sentence after Mitsubishi, Wrap-Up lets objects inherit linguistic patterns from objects to which they point. A link had been added from SOJ to DRAM, which occurs in the same sentence with Mitsubishi. The Packaging-Purchaser-Links tree returned positive for the instance with packaging-type SOJ, distance of 0, and trigger-count of 10.

A new object is created for each link between company and process, and these "microelectronics-capability" objects are then merged together according to judgments made by links-merge trees. The final output with links added is shown in figure 4.

A different example illustrates the power of Wrap-Up's links-merge stage to learn a different kind of domain knowledge. Consider the following fragment of text where a lithography process is associated with three types of chips and two pieces of equipment.

```
...a clean room utilizing Ultratech Stepper and
GCA steppers.  General Signal said it plans to
use the facility to demonstrate manufacturing
techniques, such as mix-and-match lithography
methods to produce dynamic and static RAMs and
ASIC devices.
```

In this domain, a process may point to several devices in the output, in this example lithography linked to DRAM, SRAM, and ASIC. But if a process is linked to multiple equipment, it is typically considered multiple processes, each one pointing to a separate equipment object. The target output for this text has lithography-1 pointing to Ultratech Stepper and to DRAM, SRAM, and ASIC. A separate lithography-2 object points to the GCA stepper and each of the three devices.

Wrap-Up uses links-merge trees to decide whether to merge or split when an object has pointers to multiple objects. Interestingly enough this domain knowledge could only be learned by example from the training corpus, as it was mentioned nowhere in the fifty pages of domain guidelines supplied by ARPA to MUC-5 participants.

Wrap-Up is also able to infer objects not explicitly extracted from the text and to add context-sensitive defaults for some slot values. If a text has stepper equipment, but no process using the stepper is mentioned, it becomes an "orphan" with no object pointing to it in the output. The Equipment-Orphans tree learned that stepper equipment always had a lithography process pointing to it in the training output. The "orphans" trees return the type of object to be added to the output, if a parent object can be inferred.

## 4. Test Results

The performance of Wrap-Up compared well with that of the official UMass/Hughes MUC-5 system, where output from the CIRCUS sentence analyzer was sent to TTG (Trainable Template Generator), a discourse component based on the Trainable Text Skimmer from the Hughes Research Laboratories (Dolan, et al. 1991, Lehnert et al. 1993).

Acquisition of domain knowledge by machine learning was at the heart of the TTG system, but it didn't go as far as Wrap-Up in being fully trainable. Some of the features used by TTG classifiers were generated by domain-specific code, and decisions about merging or splitting, which Wrap-Up handles during its links-merge stage, were done by hand-coded heuristics external to TTG. The ability to infer objects not explicitly mentioned in the text was also more limited than that of Wrap-Up.

Several iterations of hand-tuning were required to adjust thresholds for the decision trees produced by TTG, where Wrap-Up uses ten-fold cross-validation to automatically evaluate different thresholds and pruning levels. After a day of CPU-time building decision trees, Wrap-Up is a working system with no further programming effort needed. Additional fine-tuning can be done, but the results shown in figure 5 are with no fine-tuning.

Wrap-Up outperformed TTG in both overall recall and precision on the official MUC-5 micro-electronics test sets. Performance metrics used in the MUC evaluations are recall, precision, and f-measure. Recall is the percentage of possible information that was reported. Precision is the percent correct of the reported information. F-measure combines these into a single metric with the formula $F = ((\beta^2 + 1)PR)/(\beta^2 P + R)$, where $\beta$ is set to 1 here.

|        | Wrap-Up |       |      | TTG  |       |      |
|--------|---------|-------|------|------|-------|------|
|        | Rec.    | Prec. | F    | Rec. | Prec. | F    |
| Part 1 | 32.3    | 44.4  | 37.4 | 27.1 | 39.5  | 32.1 |
| Part 2 | 36.3    | 38.6  | 37.4 | 32.7 | 37.0  | 34.7 |
| Part 3 | 34.6    | 37.7  | 36.1 | 34.7 | 40.5  | 37.5 |
| Avg.   | 34.4    | 40.2  | 36.8 | 31.5 | 39.0  | 34.8 |

Figure 5: Performance on MUC-5 microelectronics test sets

Lack of coverage by the sentence analyzer places a ceiling on recall for the discourse component. In test set part 1 there were 208 company names to be extracted. The CIRCUS analyzer extracted a total of 404 company names, with only 131 correct and 2 partially correct, giving a baseline of 63% recall and 33% precision for that slot. Wrap-Up's entity-name filter tree managed to discard a little over half of the spurious company names, keeping 77% of the good companies. This resulted in 49% recall and 44% precision for this slot. TTG discarded a little less than half the spurious companies, but kept only 59% of the good ones, resulting in 40% recall and 36% precision for this slot.

Although precision is often increased at the expense of recall, Wrap-Up also has mechanisms to generate a small increase in recall. Inferring a lithography process from stepper equipment, or splitting a process that is linked to multiple equipment can gain back recall that is lost from discarding objects during the filter stage.

## 5. Conclusions

Information extraction systems represent a new and exciting class of applications for NLP technologies. ARPA-sponsored performance evaluations have demonstrated the importance of domain portability and fast system development cycles in making IE systems economically viable (Sundheim 1991, 1992, 1993; Lehnert and Sundheim 1991). In an effort to address these issues, researchers have show that representative text corpora can be exploited to solve problems at the level of sentence analysis (Riloff 1993, Cardie 1993, Hobbes et al. 1992, Ayuso et al. 1992). Our work shows that representative text corpora can be exploited in order to handle problems at the level of discourse analysis as well.

Our approach requires a set of hand-crafted answer keys in addition to source texts, and this resource represents a labor-intensive investment on the part of domain experts. On the other hand, no knowledge of NLP or ML technologies is needed to generate these answer keys, so any domain expert can produce answer keys for use by Wrap-Up. It is also easier to generate a few hundred answer keys than it is to write down explicit and comprehensive domain guidelines. Moreover, domain knowledge implicitly present in a set of answer keys may go beyond the conventional knowledge of a domain expert when reliable patterns of information transcend a logical domain model.

Because Wrap-Up requires no hand-coded heuristics or manual design, it provides a paradigm for user-customizable system design, where no technological background on the part of the user is assumed. We have seen how Wrap-Up improves recall and precision produced at the level of the sentence analyzer. This suggests that improvements in overall system performance can be obtained by improving the operation of the sentence analyzer, or perhaps through feedback between sentence analysis and discourse analysis.

The integration of ML algorithms in a comprehensive IE system also encourages a new perspective on sentence analysis. If ML technologies are especially successful at noise reduction, it makes sense to pursue sentence analysis techniques that favor recall over precision. While we normally expect error rates to propagate across a serial system, noise-tolerant ML algorithms may be able to hold error rates in check as we move from low levels of text analysis to the highest levels of language comprehension.

But even if ML algorithms could only duplicate the performance levels of hand-coded discourse modules, we would still be looking at a major achievement in terms of portability and scalability. Our experience with Wrap-Up suggests that ML algorithms provide a promising foundation for corpus-driven discourse analysis. Hand-coded heuristics can be replaced by decision trees, and implicit domain knowledge can be derived from a representative development corpus. This result is both encouraging with respect to practical system development, and somewhat provocative with respect to the larger issue of automated knowledge acquisition.